\documentclass{article}

\usepackage{microtype}
\usepackage{graphicx}
\usepackage{subfigure}
\usepackage{booktabs} 

\usepackage{hyperref}

\usepackage[accepted]{icml2019}

\usepackage[utf8]{inputenc} 
\usepackage[T1]{fontenc}    
\usepackage{url}            
\usepackage{booktabs}       
\usepackage{amsfonts}       
\usepackage{nicefrac}       
\usepackage[scaled=0.8]{beramono}
\usepackage{amsmath}
\usepackage{amssymb}
\usepackage{xcolor,colortbl}
\usepackage{listings}
\usepackage{paralist}
\usepackage{wrapfig}
\usepackage{enumitem}
\usepackage{multicol}
\usepackage{flushend}
\usepackage{nicefrac}
\usepackage{xfrac}
\usepackage{tikz}
\usetikzlibrary{positioning}
\definecolor{processblue}{cmyk}{0.96,0,0,0}

\lstdefinelanguage{DOT}%
{morekeywords={val,new},%
  sensitive,%
  morecomment=[l]//,%
  morecomment=[s]{/*}{*/},%
  morestring=[b]",%
  morestring=[b]',%
  showstringspaces=false%
}[keywords,comments,strings]%

\newlength{\trulemargin}
\newlength{\trulewidth}
\newlength{\srulewidth}
\newenvironment{trules}{$\vspace{0.5em}\ba{p{\trulemargin}@{~}p{\trulewidth}@{~}p{\trulemargin}}}{\ea$}
\newenvironment{srules}{$\vspace{0.5em}\ba{p{\trulemargin}@{~}p{\srulewidth}}}{\ea$}

\newcommand{\ba}{\begin{array}}
\newcommand{\ea}{\end{array}}

\newcommand{\ei}{\end{array}}
\newcommand{\bcases}{\left\{\begin{array}{ll}}
\newcommand{\ecases}{\end{array}\right.}

\graphicspath{{figures/}}

\icmltitlerunning{Graph Neural Reasoning for 2-Quantified Boolean Formula Solver}

\begin{document}

\twocolumn[
\icmltitle{Graph Neural Reasoning for 2-Quantified Boolean Formula Solvers}



\icmlsetsymbol{equal}{*}

\begin{icmlauthorlist}
\icmlauthor{Zhanfu Yang*}{to}
\icmlauthor{Fei Wang*}{to}
\icmlauthor{Ziliang Chen}{yo}
\icmlauthor{Guannan Wei}{to}
\icmlauthor{Tiark Rompf}{to}
\end{icmlauthorlist}

\icmlaffiliation{to}{Department of Computer Science, Purdue University, USA}
\icmlaffiliation{yo}{Sun Yat-Sen University, China}

\icmlcorrespondingauthor{Zhanfu Yang}{yang1676@purdue.edu}


\vskip 0.3in
]



\printAffiliationsAndNotice{}  

\lstMakeShortInline[keywordstyle=,%
              flexiblecolumns=false,%
              mathescape=false,%
              basicstyle=\tt]@

\begin{abstract}
In this paper, we investigate the feasibility of learning GNN (Graph Neural
Network) based solvers and GNN-based heuristics for specified QBF (Quantified Boolean
Formula) problems. We design and evaluate several GNN architectures for 2QBF
formulae, and conjecture that GNN has limitations
in learning 2QBF solvers. Then we show
how to learn a heuristic CEGAR 2QBF solver. We further explore
generalizing GNN-based heuristics to larger unseen instances, and uncover
some interesting challenges. In summary, this paper provides a comprehensive surveying view of applying GNN-embeddings
to specified QBF solvers, and aims to offer guidance in applying ML to more complicated
symbolic reasoning problems.
\end{abstract}
\section{Introduction}

A propositional formula expression consists of Boolean constants ($\top$:
true, $\bot$: false), Boolean variables ($x_i$), and propositional connectives
such as $\land$, $\lor$, $\lnot$, and etc. The SAT (Boolean Satisfiability)
problem, which asks if given a formula can be satisfied (as $\top$) by
assigning proper Boolean values to the variables, is the first proven
NP-complete problem \cite{Cook:1971:CTP:800157.805047}.  As an extension of propositional formula, QBF (Quantified Boolean Formula) allows quantifiers ($\forall$ and $\exists$) over the Boolean variables. In general, a quantified Boolean formula can be expressed as such:
\vspace{-1.5ex}
$$
Q_i X_i Q_{i-1} X_{i-1} ... Q_0 X_0 \phi
\vspace{-1.5ex}
$$
where $Q_i$ denote quantifiers that differ from its neighboring
quantifiers, $X_i$ are disjoint sets of variables, and $\phi$
is propositional formulae with all Boolean variables bounded. The QBF problem is PSPACE-complete
\cite{SAVITCH1970177}. To this researchers previously proposed incremental determinzation
\citep{DBLP:conf/sat/RabeS16, 10.1007/978-3-319-96142-2_17} or CEGAR-based \citep{DBLP:journals/ai/JanotaKMC16} solvers to solve it.
They are non-deterministic, \emph{e.g.}, employing
heuristics guidance for search a solution.
Recently, MaxSAT-based
\cite{DBLP:conf/sat/JanotaS11} and ML-based \citep{DBLP:conf/aaai/Janota18} heuristics have
been proposed into CEGAR-based solvers. Without existing decision procedure, \citet{DBLP:journals/corr/abs-1802-03685} presented a GNN architecture that
embeds the propositional formulae. \citet{amizadeh2018learning} adapt a RL-style
explore-exploit mechanism in this problem, but considering circuit-SAT problems.
However, these solvers didn't tackle unsatisfiable formulae. In terms of above discussion, there are no desirable general solver towards a QBF problem in practice. To this end, we focus on 2QBF formulae in this paper, a specified-QBF case with only 1 alternation of quantifiers.

Extended from SAT, 2QBF problems keep attracting a lot of attentions due to their practical usages \cite{DBLP:conf/fpga/MishchenkoBFG15, DBLP:conf/sat/MneimnehS03, DBLP:journals/jar/RemshagenT05}, yet remaining very challenging like QBF.  Formally, $ Q_1 X Q_2 Y . \phi$, where $Q_i \in \{ \forall ,\exists \}$, $X$
and $Y$ are sets of variables, and $\phi$ is quantifier-free formula.
The quantifier-free formula $\phi$ can be in Conjunctive
Normal Form (CNF), where $\phi$ is a conjunction of \textit{clauses},
clauses are disjunctions of \textit{literals},
and each literal is either a variable or its negation. For
example, the following term is a well-formed 2QBF in CNF:
$ \forall x, y \exists z . (x \lor z) \land (y \lor \neg z) $.
If $\phi$ is in CNF, it is required that the $\forall$ quantifier is
on the outside, and the $\exists$ quantifier is on the inside. Briefly, the 2QBF problem is to ask whether
the formula can be evaluated to $\top$ considering the $\forall$ and $\exists$
quantifications. It's presumably exponentially harder to solve 2QBF than SAT because it characterizes the second level of the polynomial hierarchy. 

Our work explores several different 2QBF solvers by way of graph neural-symbolic
reasoning. In Section 2, we investigate famous SAT GNN-based solvers \cite{DBLP:journals/corr/abs-1802-03685}\cite{amizadeh2018learning}. We found these architectures hard to extend to 2QBF problems, due to that GNN is
unable to reason about unsatisfiability. To this, we further make some effective reconfiguration to GNN. In Section 3, on behalf of a traditional CEGAR-based solver, three ways to learn the GNN-based heuristics are proposed: to rank the candidates, to rank the counterexamples, and their combination. They aim to avoid
multiple GNN embeddings per formula, to reduce the GNN inference
overhead. Relevant experiments showcase their superiorities in 2QBF.

\section{GNN-based QBF Solver Failed}\label{sec:learn_solver}\vspace{-6pt}
Let's first revisit the existing GNN-based SAT solvers, and analyze why they fails to suit the 2QBF problem. 

\subsection{GNN for QBF}
\vspace{-1ex}
\paragraph{Embedding of SAT}
SAT formulae are translated into bipartite graphs \citet{DBLP:journals/corr/abs-1802-03685}, where literals ($L$) represent one kind of nodes,
and clauses ($C$) represent the other kind. We denote EdgeMatrix ($\mathbb{E}$) as edges between literal and clause nodes with dimension $|C|$ x $|L|$.
The graph of $(x \lor \neg y) \land (\neg x \lor y)$ is given below as an example.
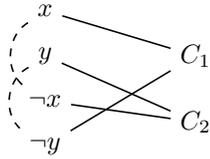
\vspace{-2ex}
\begin{center}

\begin{tikzpicture}[-latex, auto, node distance=0.6cm and 2.0cm, on grid, semithick,
  state/.style ={
draw=none, text=black, minimum width = 0.3cm}]
\node[state] (A) {$C_1$};
\node[state] (B) [below = of A, yshift=-8pt] {$C_2$};
\node[state] (C) [above left = of A] {$x$};
\node[state] (D) [left = of A] {$y$};
\node[state] (E) [below left = of A] {$\neg x$};
\node[state] (F) [below = of E] {$\neg y$};
\draw[-]
  (A) edge (C) (A) edge (F) (B) edge (D) (B) edge (E);
\draw[-,dashed, bend right = 55]
  (C) edge (E) (D) edge (F);
\end{tikzpicture}
\end{center}
\vspace{-3ex}

As below, $\text{Emb}_L$ and $\text{Emb}_C$ denote embedding matrices of literals and clauses respectively,
$\text{Msg}_{X \to Y}$ denotes messages from $X$ to $Y$,
$\mathcal{M}_X$ denotes MLP of $X$ for generating messages,
$\mathcal{L}_X$ denotes LSTM of $X$ for digesting incoming messages and updating embeddings,
$X \cdot Y$ denotes matrix multiplication of $X$ and $Y$,
$X^T$ denotes matrix transportation of $X$,
$[X, Y]$ denotes matrix concatenation, and 
$\text{Emb}_{\neg L}$ denotes the embedding of $L$'s negations.

\vspace{-3ex}
$$\small
\ba{cl}
\text{Msg}_{L \to C} = \mathcal{M}_L (\text{Emb}_L) \\
\text{Emb}_{C} = \mathcal{L}_C (\mathbb{E} \cdot \text{Msg}_{L \to C}) \\
\text{Msg}_{C \to L} = \mathcal{M}_C (\text{Emb}_C) \\
\text{Emb}_L = \mathcal{L}_L ([\mathbb{E}^T \cdot \text{Msg}_{C \to L}, \text{Emb}_{\neg L}]) 
\ea
$$
\vspace{-3ex}

Iterations are fixed for train but can be unbounded for
test.
\vspace{-2ex}
\paragraph{Embedding of 2QBF}

We separate $\forall$-literals and
$\exists$-literals in different groups, embed them via different NN modules.
The graph representation of $\forall x \exists y . (x \lor \neg y) \land (\neg x \lor y)$
shows:

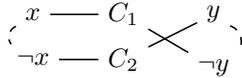
\vspace{-2.5ex}
\begin{center}
\begin{tikzpicture}[-latex, auto, node distance=0.6cm and 1.2cm, on grid, semithick,
state/.style ={
draw=none, text=black, minimum width = 0.2cm}]
\node[state] (A) {$C_1$};
\node[state] (B) [below = of A] {$C_2$};
\node[state] (C) [left = of A] {$x$};
\node[state] (D) [left = of B] {$\neg x$};
\node[state] (E) [right = of A] {$y$};
\node[state] (F) [right = of B, yshift=-3pt] {$\neg y$};
\draw[-]
  (A) edge (C) (A) edge (F) (B) edge (D) (B) edge (E);
\draw[-,dashed, bend right = 55]
  (C) edge (D);
\draw[-,dashed, bend left = 55]
  (E) edge (F);
\end{tikzpicture}
\end{center}
\vspace{-3ex}

We use $\forall$ and $\exists$ to denote all $\forall$-literals and all $\exists$-literals respectively.
We use $\mathbb{E}_X$ denote the EdgeMatrix between $X$ and $C$, and
$\mathcal{M}_{C\to X}$ denote MLPs that generate $\text{Msg}_{C\to X}$.

\vspace{-3ex}
$$\small
\ba{cl}
\text{Msg}_{\forall\to C} = \mathcal{M}_\forall (\text{Emb}_\forall) \\
\text{Msg}_{\exists\to C} = \mathcal{M}_\exists (\text{Emb}_\exists) \\
\text{Emb}_{C} = \mathcal{L}_{C} ([\mathbb{E}_\forall \cdot \text{Msg}_{\forall\to C}, \mathbb{E}_\exists \cdot \text{Msg}_{\exists\to C}]) \\
\text{Msg}_{C\to\forall} = \mathcal{M}_{C\to\forall} (\text{Emb}_{C}) \\
\text{Msg}_{C\to\exists} = \mathcal{M}_{C\to\exists} (\text{Emb}_{C}) \\
\text{Emb}_\forall = \mathcal{L}_\forall ([\mathbb{E}_\forall^T
 \cdot \text{Msg}_{C\to\forall}, \text{Emb}_{\neg \forall}]) \\
\text{Emb}_\exists = \mathcal{L}_\exists ([\mathbb{E}_\exists^T
\cdot \text{Msg}_{C\to\exists}, \text{Emb}_{\neg \exists}]) \\
\ea
\vspace{-1ex}
$$
We designed multiple architectures (details in supplementary) and use the best one as above for the rest of the paper.
\vspace{-4ex}
\paragraph{Data Preparation}
For training and testing, we follow \citet{DBLP:conf/ijcai/ChenI05}, which generates
QBFs in conjunctive normal form. Specifically, we generate problems of \emph{specs} (2,3) and \emph{sizes}
(8,10). Each clause has 5 literals, 2 of them are randomly chosen from
a set of 8 $\forall$-quantified variables, 3 are randomly chosen from a set
of 10 $\exists$-quantified variables. We modify the generation
procedure that it generates clauses until the formula becomes
unsatisfiable. We then randomly negate an $\exists$-quantified literal per formula to make
it satisfiable. 
\vspace{-2ex}
\paragraph{SAT/UNSAT}

We vote MLPs from $\forall$-variables and use average votes as logits for SAT/UNSAT prediction:
\vspace{-2ex}
$$\text{logits} = \text{mean}(\mathcal{M}_{vote}(\text{Emb}_\forall))
\vspace{-3ex}$$

As in table~\ref{table1}, Each block of entries are
accuracy rate of UNSAT and SAT formulae respectively. The models are tested on
600 pairs of formulae and we
allow message-passing iterations up to 1000. GNNs fit well to smaller training dataset,
but has trouble for 160 pairs of formulae. Performance
deteriorates when embedding iterations increase and most GNNs become
very biased at high iterations. 
\begin{table}[t]
\caption{GNN Performance to Predict SAT/UNSAT}
\label{table1}
\vskip 0.05in
\begin{center}
\begin{small}
\begin{sc}
\begin{tabular}{lcccr}
\toprule
Dataset    & 40 pairs & 80 pairs & 160 pairs \\
\midrule
8 iters    & (0.98, 0.94) & (1.00, 0.92) & (0.84, 0.76) \\
testing    & (0.40, 0.64) & (0.50, 0.48) & (0.50, 0.50) \\
\midrule
16 iters   & (1.00, 1.00) & (0.96, 0.96) & (0.88, 0.70) \\
testing    & (0.54, 0.46) & (0.52, 0.52) & (0.54, 0.48) \\
\midrule
32 iters   & (1.00, 1.00) & (0.98, 0.98) & (0.84, 0.80) \\
testing    & (0.32, 0.68) & (0.52, 0.50) & (0.52, 0.50) \\
\bottomrule
\end{tabular}
\end{sc}
\end{small}
\end{center}
\vskip -0.2in
\end{table}

\begin{table}[t]
\caption{GNN Performance to Predict Witness of UNSAT}
\label{table2}
\vskip 0.05in
\begin{center}
\begin{small}
\begin{sc}
\begin{tabular}{lcccr}
\toprule
Dataset    & 160 unsat & 320 unsat & 640 unsat \\
\midrule
8 iters    & (1.00, 0.99) & (0.95, 0.72) & (0.82, 0.28) \\
testing    & (0.64, 0.06) & (0.67, 0.05) & (0.69, 0.05) \\ 
\midrule
16 iters   & (1.00, 1.00) & (0.98, 0.87) & (0.95, 0.69) \\
testing    & (0.64, 0.05) & (0.65, 0.05) & (0.65, 0.06) \\
\midrule
32 iters   & (1.00, 1.00) & (0.99, 0.96) & (0.91, 0.57) \\
testing    & (0.63, 0.05) & (0.64, 0.05) & (0.63, 0.05) \\  
\bottomrule
\end{tabular} 
\end{sc}
\end{small}
\end{center}  
\vskip -0.2in
\end{table}

\vspace{-2ex}
\paragraph{$\forall$-Witnesses of UNSAT} \label{predict_candidate}
Proving unsatisfiability of 2QBF needs a witness of unsatisfiability,
which is an assignment to $\forall$-variables that eventually leads to
UNSAT. We use logistic regression in this experiment. To be specific, the final embeddings of $\forall$-variables
are transformed into logits via a MLP $\mathcal{M}_{asn}$ and used to compute the
cross-entropy loss with the known witness unsatisfiability of the formulae.
\vspace{-1.5ex}
$$\text{witness} = \text{softmax}(\mathcal{M}_{asn}(\text{Emb}_\forall))\vspace{-1ex}$$
This training task is very similar to 
\citet{amizadeh2018learning}, except our GNN has to
reason about unsatisfiability of the simplified SAT formulae, which we believe
infeasible. We summarize the results in Table ~\ref{table2}.
In each block of entries, we list the accuracy per
variable and accuracy per formulae on the left and right seperately. Entries in upper
half of each block is for training data, and lower half for testing data. 
From the table we see that GNNs fit well to the training
data. More iterations of message-passing give better fitting. However,
the performance on testing data are only
slightly better than random. More iterations in testing do not help
with performance.

\subsection{Why GNN-based QBF Solver Failed} \label{sec:embedding}
We conjecture current GNN architectures and embedding processes are unlikely to prove unsatisfiability or reason about
$\forall$-assignments. Even in SAT problem
\citet{DBLP:journals/corr/abs-1802-03685}, GNNs are good at finding solutions
for satisfiable formulae, while not for confidently proving
unsatisfiability. Similarly
\citet{amizadeh2018learning} had little success in proving unsatisfiability with
DAG-embedding because showing SAT only needs a witness, but proving UNSAT needs more complete reasoning about the search
space. A DPLL-based approach would iterate all possible assignments and
construct a proof of UNSAT. However,
a GNN embedding process is neither following a strict order of assignments, nor
learning new knowledge that indicates some assignments should be avoided. In fact, the GNN embedding may be mostly similar to vanilla WalkSAT approaches,
with randomly initialized assignments and stochastic local search, which can not prove unsatisfiability.

This conjecture may be a great obstacle for learning 2QBF solvers
from GNN, because proving either satisfiability or unsatisfiability of the 2QBF
problem needs not only a witness. If the formula is satisfiable, proof needs
to provide assignments to $\exists$-variables under all possible assignments of
$\forall$-variables or in a
CEGAR-based solver. If the
formula is unsatisfiable, then the procedure should find an assignment for the
$\forall$-variables. 

\section{Learn GNN-based Heuristics}\label{sec:gnn_heuristics}
In Section ~\ref{sec:learn_solver}, we know that GNN-based 2QBF
solvers are unlikely to be learned, therefore, the success of learning SAT solvers
\citep{DBLP:journals/corr/abs-1802-03685, amizadeh2018learning} cannot simply
extend to 2QBF or more expressive logic. We consider the
CEGAR-based solving algorithm, to reduce
the GNN inference overhead. We first present the CEGAR-based solving procedure in Algorithm \ref{alg:2qbf}
\cite{DBLP:conf/sat/JanotaS11}.
\begin{algorithm}
\small
   \caption{CEGAR 2QBF solver}
   \label{alg:2qbf}
\begin{algorithmic}
   \STATE {\bfseries Input:} $\forall X \exists Y \phi$
   \STATE {\bfseries Output:} (sat, -) or (unsat, witness)
   \STATE Initialize constraints $\omega$ as empty set. \\
   \WHILE {true}
     \STATE (has-candidate, candidate) = SAT-solver($\omega$)
     \IF {not has-candidate}
        \STATE \textbf {return} (sat, -)
     \ENDIF
     \STATE (has-counter, counter) = SAT-solver($\phi[X \rightarrow \text{candidate}]$)
     \IF {not has-counter}
        \STATE \textbf {return} (unsat, candidate)
     \ENDIF
     \STATE add counter to constraints $\omega$
   \ENDWHILE 
\end{algorithmic}
\end{algorithm}
Note that $\omega$ is constraints for candidates.
Initially, $\omega$ is $\emptyset$, and any assignment of $\forall$-variables can be proposed as candidate 
which may reduce the problem to a smaller propositional formula.
If we can find an assignment to $\exists$-variables that satisfies the propositional formula,
this assignment is called a counterexample to the candidate.
We denote $\phi_{counter}$ as all clauses in $\phi$ that are satisfied by the counterexample.
The counterexample can be transformed into a constraint, stating that next candidates
cannot simultaneously satisfy clauses ($\phi \setminus \phi_{counter}$),
since those candidates are already rejected by the current counterexample.
This constraint can be added to $\omega$ as a propositional term,
thus finding new candidates is done by solving constraints-derived propositional term $\omega$.
\subsection{Ranking the Candidates}
In order to decide which candidate to use from $\text{SAT-solver}$ $(\omega)$,
we can rank solutions in MaxSAT-style by simplifying the
formula with candidates and ranking them based on the number of clauses they satisfy.
We use it as a benchmark comparison. 
Besides, the hardness can be evaluated as the number of solutions of the simplified
propositional formula. Thus the training data of our ranking GNN is
all possible assignments of $\forall$-variables
and the ranking scores that negatively relate to the number of solutions of each
assignment-propagated propositional formula (Details about computing the ranking scores
shown in supplementary).

We extend the GNN embedding architecture so that the final embedding of the
$\forall$-variables are transformed into a scoring matrix ($\text{Sm}_\forall$)
for candidates via a MLP ($\mathcal{M}_{\forall, scoring}$). A batch of candidates
($\mathbb{C}$) are ranked by passing through a two-layer MLP without biases,
where the weights of the first layer is the scoring matrix
($\text{Sm}_\forall$), and the weights of the second layer is a weight vector
($\text{Wv}_\forall$).
\vspace{-1ex}
$$\small
\begin{array}{cl}
\text{Sm}_\forall = \mathcal{M}_{\forall,\text{scoring}} (\text{Emb}_\forall) \\
\text{Score}_\forall = \text{ReLU}(\mathbb{C} \cdot \text{Sm}_\forall) \cdot \text{Wv}_\forall\\
\end{array}
\vspace{-1.5ex}
$$
We make use of the TensorFlow ranking library \cite{DBLP:journals/corr/abs-1812-00073}
to compute the pairwise-logistic-loss with NDCG-lambda-weight for supervised training. 
What's more, we evaluate our ranking heuristics by adding them to CEGAR cycle and measure the average steps needed to solve the problems. It requires us to change the $SAT(\omega)$ subroutine to a $nSAT(\omega)$ subroutine, where
once a solution is found, it is added back to the formula as constraint, and
search for a different solution, until no solutions can be found or maximal
number of solutions is reached. Then the heuristics ranks the solutions and proposes the best one as candidate.
We use 4 datasets: (1)TrainU: 1000 unsatisfiable formulae used for training;
(2) TrainS: 1000 satisfiable formulae used for training;
(3) TestU: 600 unsatisfiable formulae used for testing;
(4) TestS: 600 satisfiable formulae used for testing); and 4 ranking heuristics:
(1) -: no ranking;
(2) MaxSAT: ranking by the number of satisfied clauses via on-the-fly formula
simplification;
(3) GNN1: ranking by hardness via GNN model inference;
(4) GNN2: ranking by the number of satisfied clauses via GNN model inference.

As shown in Table ~\ref{table4}, all 3 ranking heuristics improve the solving process of all 4 datasets.
Unsatisfiable formulae benefit more from the heuristics, and the heuristics generalizes
very well from training formulae to testing formulae. Machine learning results are repeated twice with different random seeds, and
numbers shown are from models with best performance on training data.

\begin{table}[t]
\caption{Performance of CEGAR Candidate Ranking}
\label{table4}
\vskip 0.05in
\begin{center}
\begin{small}
\begin{sc}
\begin{tabular}{lcccr}
\toprule
DataSet   & TrainU & TrainS & TestU & TestS \\
\midrule
-         & 21.976 & 34.783 & 21.945 & 33.885  \\
maxSAT    & 13.144 & 30.057 & 12.453 & 28.863  \\
\midrule
GNN1      & 13.843 & 31.704 & 13.988 & 30.573  \\
GNN2      & 15.287 & 32.0   & 14.473 & 30.788  \\
\bottomrule
\end{tabular}
\end{sc}
\end{small}
\end{center}
\vskip -0.35in
\end{table}

\subsection{Ranking the Counterexamples}
We consider a GNN-based heuristics for ranking counterexamples. Each
counterexample contributes to a constraint in $\omega$, which either shrinks the
search space of the witnesses of unsatisfiability, or be added to the
constraints indicating that no candidates are witnesses of unsatisfiability.

As following, we compute ranking scores for our training data.
For satisfiable 2QBF instances in the training data, we list all possible assignments of $\exists$-variables,
and collect all constraining clauses in $\omega$.
Then we solve $\omega$ with \texttt{hmucSAT} \cite{DBLP:conf/fmcad/NadelRS13},
seeking for unsatisfiability cores.
Initially we plan to give a high ranking score (10) for $\exists$-assignments
corresponding to clauses in unsatisfiability cores, and a low ranking score (1)
for all other $\exists$-assignments.
Later, we choose to give other $\exists$-assignments ranking scores based on the number of satisfied clauses,
in range of $[1, 8]$ because unsatisfiability cores are often small.

For unsatisfiable 2QBF instances, we collect all
constraining clauses in $\omega$. As $\omega$ is satisfiable 
and solutions are actually witnesses of unsatisfiability. To obtain unsatisfiability
scores, we add solutions to $\omega$ as extra constraints until the
$\omega$ becomes unsatisfiable. We then compute the ranking scores.

We use another dataset of which rankings scores is totally based on
the number of clauses satisfied for comparison. To add ranking modules, we extend GNN embedding architecture. Notations include $\text{Sm}_\exists$ for the
scoring matrix, $\mathcal{M}_{\exists, \text{scoring}}$ for a MLP to get scoring
matrix from the final embedding of $\exists$-variables, $\mathbb{CE}$ for a batch of
counterexamples, and $\text{Wv}_\exists$ for the weight vector.

\vspace{-2.8ex}
$$\small
\begin{array}{cl}
\text{Sm}_\exists = \mathcal{M}_{\exists, \text{scoring}} (\text{Emb}_\exists) \\
\text{Score}_\exists = \text{ReLU}(\mathbb{CE} \cdot \text{Sm}_\exists) \cdot \text{Wv}_\exists
\end{array}
\vspace{-2.8ex}
$$

\begin{table}[t]
\caption{Performance of CEGAR-COUNTER-RANKING}
\label{table5}
\vskip 0.05in
\begin{center}
\begin{small}
\begin{sc}
\begin{tabular}{lcccr}
\toprule
DataSet   & TrainU & TrainS & TestU & TestS \\
\midrule
-         & 21.976 & 34.783 & 21.945 & 33.885  \\
maxSAT    & 14.754 & 22.265 & 14.748 & 21.638  \\
\midrule
GNN1      & 17.492 & 26.962 & 17.198 & 26.598  \\
GNN2      & 16.95  & 26.717 & 16.743 & 26.325  \\
\bottomrule
\end{tabular}
\end{sc}
\end{small}
\end{center}
\vskip -0.25in
\end{table}

After supervised training, we evaluate the trained GNN-based ranking heuristics
in a CEGAR-based solver. The results are shown in Table
~\ref{table5}. Based on the MaxSAT heuristics, ranking counterexamples benefits
solving satisfiable formulae more than unsatisfiable formulae. However, GNN1  performs worse than GNN2. The likely explanation is that
predicting unsatisfiability cores is far too complicated for GNN. Moreover,
knowledge of unsatisfiability cores cannot be obtained from each counterexample
alone, but needs analysis of all counterexamples collectively. It may go back
to the limitation of GNN in reasoning about ``all possible solutions'', and the
added score information behaves like an interference rather than knowledge for
GNN-based ranking heuristics. Machine learning results are repeated twice,
reporting models with best training data performance.
\begin{table}[t]
\caption{Performance of CEGAR-BOTH-RANKING}

\label{table6}
\vskip 0.05in
\begin{center}
\begin{small}
\begin{sc}
\begin{tabular}{lcccr}
\toprule
DataSet   & TrainU & TrainS & TestU & TestS \\
\midrule
-         & 21.976 & 34.783 & 21.945 & 33.885  \\
maxSAT    & 9.671  & 20.777 & 9.425  & 19.883  \\
\midrule
GNN1      & 11.686 & 25.021 & 11.605 & 24.518  \\
GNN2      & 12.505 & 25.505 & 12.22  & 24.938  \\
GNN3      & 11.25  & 24.76  & 12.008 & 24.295  \\
\bottomrule
\end{tabular}
\end{sc}
\end{small}
\end{center}
\vskip -0.25in
\end{table}

\subsection{Combination of the Heuristics}
To combine ranking heuristics and counterexamples in a
single solver, we extend the GNN-embedding architecture with ranking data of candidates
and counterexamples. We have GNN1 trained by ranking scores from hardness and
unsatisfiability cores, GNN2 trained by ranking scores from the number of
satisfied clauses for both candidates and counterexamples, and GNN3 trained by
ranking scores from hardness for candidates, and number of satisfied clauses for
counterexamples. As in Table ~\ref{table6}, GNN3 is arguably the best model we obtained from supervised learning via this ranking method.
All machine learning results are repeated twice with different random seeds,
and models with best performance in training data are reported.

\section{Conclusion}\label{sec:conclusion}
In this paper, we show learning GNN-based 2QBF solvers is hard by
current GNN architectures due to its inability to reason about unsatisfiability. Our work extends the previous GNN-based 2QBF solver in terms of CEGAR-based heuristic. A suite of GNN-based techniques have been made
to improve the GNN embedding for reasoning 2QBF solutions. Their superiorities are witnessed in our experiments.
\bibliography{references}
\bibliographystyle{icml2019}

\end{document}